# Adversarial Examples and the Deeper Riddle of Induction:
## The Need for a Theory of Artifacts in Deep Learning

**Cameron Buckner**
Associate Professor
The University of Houston
cjbuckner@uh.edu

**Abstract:** Deep learning is currently the most widespread and successful technology in artificial intelligence. It promises to push the frontier of scientific discovery beyond current limits—it is currently the state-of-the-art method for applications as diverse as interpreting fMRI data, detecting exoplanets around distant stars, modeling interactions in high-energy physics, and predicting protein folds. However, skeptics have worried that deep neural networks are black boxes whose processing is unintelligible to humans, and have called into question whether these advances can really be deemed scientific progress if humans cannot understand them. Relatedly, these systems also possess bewildering new vulnerabilities: most notably a susceptibility to "adversarial examples", which seem to cause them to spectacularly misclassify manipulated images that do not fool humans. However, adversarial examples remain mysterious even to their creators, and many of their properties have already been misunderstood. In this paper, I introduce adversarial examples and argue that they will likely become a flashpoint of debate in philosophy and diverse sciences over the next decade. Specifically, new empirical findings concerning adversarial examples have challenged the consensus view that the networks' verdicts on these cases are caused by overfitting idiosyncratic noise in the training set, and may instead be the result of detecting predictively useful "intrinsic features of the data geometry" that humans cannot perceive (Ilyas et al., 2019).

These results should cause us to re-examine classic responses to one of the deepest puzzles at the intersection of philosophy and science: Nelson Goodman's "new riddle" of induction. Specifically, they raise the possibility that progress in a number of sciences will depend upon the detection and manipulation of useful features that humans find inscrutable. Before we can evaluate this possibility, however, we must decide which (if any) of these inscrutable features are real but available only to "alien" perception and cognition, and which are distinctive artifacts of deep learning—for artifacts like lens flares or the Gibbs phenomenon can be similarly useful for prediction, but are usually seen as obstacles to scientific theorizing. Thus, machine learning researchers urgently need to develop a systematic theory of artifacts for deep neural networks, and I conclude by sketching some initial directions for this area of research.

## I.    Introduction

Deep learning neural networks have become one of the most important technologies in artificial intelligence, neuroscience, and psychology. In AI, deep learning networks are often said to be capable of superhuman performance, achieving new benchmark scores on standard tests of image recognition almost weekly. They have defeated human grandmasters in Go, a game so complex that it was once thought to be beyond the reach of AI, because there are more possibilities for an algorithm to explore in a single game than there are atoms in the universe (Silver et al., 2017). Deep learning has detected new exoplanets orbiting stars thousands of light years away from Earth (Shallue & Vanderburg, 2018); it is one of the most promising methods capable of analyzing hundreds of petabytes of data CERN generates in its attempt to validate the Standard Model in physics (Albertsson et al., 2018); and on its first attempt, the deep learning system

AlphaFold won the Critical Assessment of protein Structure Prediction (CASP) competition, predicting protein folding outcomes 15% more accurately than human scientists who had devoted their entire professional lives to this kind of task (AlQuraishi, 2019).

When their parameters are turned down to approximate mortal ranges, deep learning networks have also demonstrated great promise as models of human perception and categorization in psychology and cognitive neuroscience. With a structure inspired by anatomical discoveries in mammalian perceptual cortex (Fukushima, 1979; Hubel & Wiesel, 1967), they are regarded as the best computational models of object recognition and perceptual categorization judgments in primates (Rajalingham et al., 2018). Neuroscientists have compared the activity patterns in various intermediate layers in a deep learning network's hierarchy to firing patterns recorded from implanted electrophysiology arrays in various regions of monkey ventral stream visual cortex; both the networks and the monkeys seem to recover the same kinds of features at comparable depths of their processing hierarchies (Guest & Love, 2019; Hong et al., 2016; Kriegeskorte, 2015; Yamins & DiCarlo, 2016). There has thus been hope that not only do these models replicate the functional input-output patterns observed in primate object recognition and perceptual similarity judgments, but that they do so by modeling the hierarchical abstraction algorithms implemented by ventral stream perceptual processing in the primate brain (Buckner, 2018).

Deep learning engineers grant their models these powers by christening them in torrents of big data, which—like Achilles' dunk in the Styx—seems to necessarily leave them with a critical vulnerability. Specifically, while deep learning models can routinely achieve superior performance on novel natural data points which are similar to those they encountered in their training set, presenting them with unusual points in data space—discovered by further "adversarial" machine learning methods designed to fool deep learning systems—can cause them to produce behavior that the model evaluates with extreme confidence, but which can look to human observers looks like a bizarre mistake (Goodfellow et al., 2015). A picture of a panda modified in a way which is imperceptible to humans is suddenly labeled as containing a gibbon (with "99.3% confidence"—Fig. 1); automated vehicles blow through carefully-vandalized stop signs their recognition systems classify as yield signs (Fig. 2); and a 40-year-old white male—when accessorized with some "adversarial glasses"—is recognized by state-of-the-art facial detection software as the actress Milla Jovovich 87% of the time (Eykholt et al., 2017; I. J. Goodfellow et al., 2015; Sharif et al., 2016). These findings have curbed the enthusiasm with which many researchers regard deep learning networks, suggesting that the intelligence they exhibit is, if anything, "alien" and "brittle". Despite appearing to derive highly-structured, sophisticated knowledge from their vast amounts of training, the discoverers of adversarial examples worry that they instead merely construct "a Potemkin village that works well on naturally occurring data, but is exposed as fake when one visits points in space that do not have a high probability" (Goodfellow, Shlens, & Szegedy, 2015)."

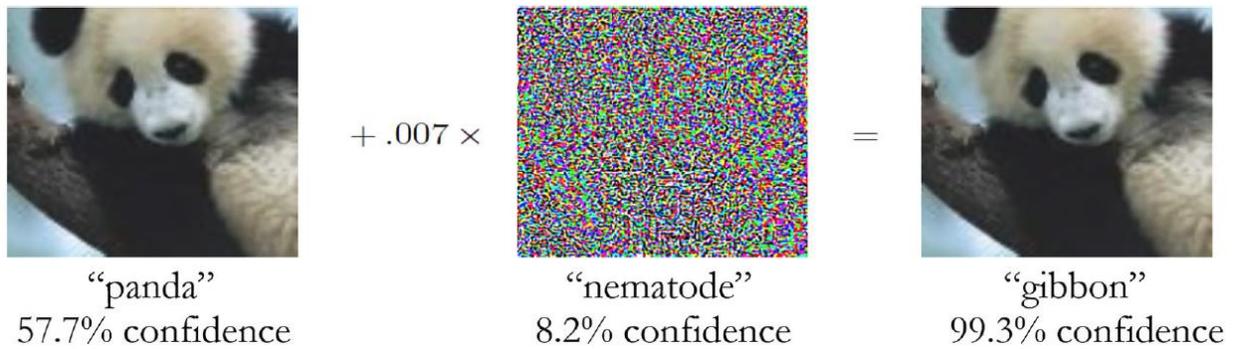

Figure 1. An adversarial "perturbed image" from the researchers that originally discovered the phenomenon, Goodfellow, Schlens, & Szegedy (2014). After the "panda" image on the left was modified with the addition of a small pixel vector (center—itself originally classified with low confidence as a nematode), the system classified the modified image on the right as a gibbon, with 99.3% confidence—despite the modification being undetectable to humans.

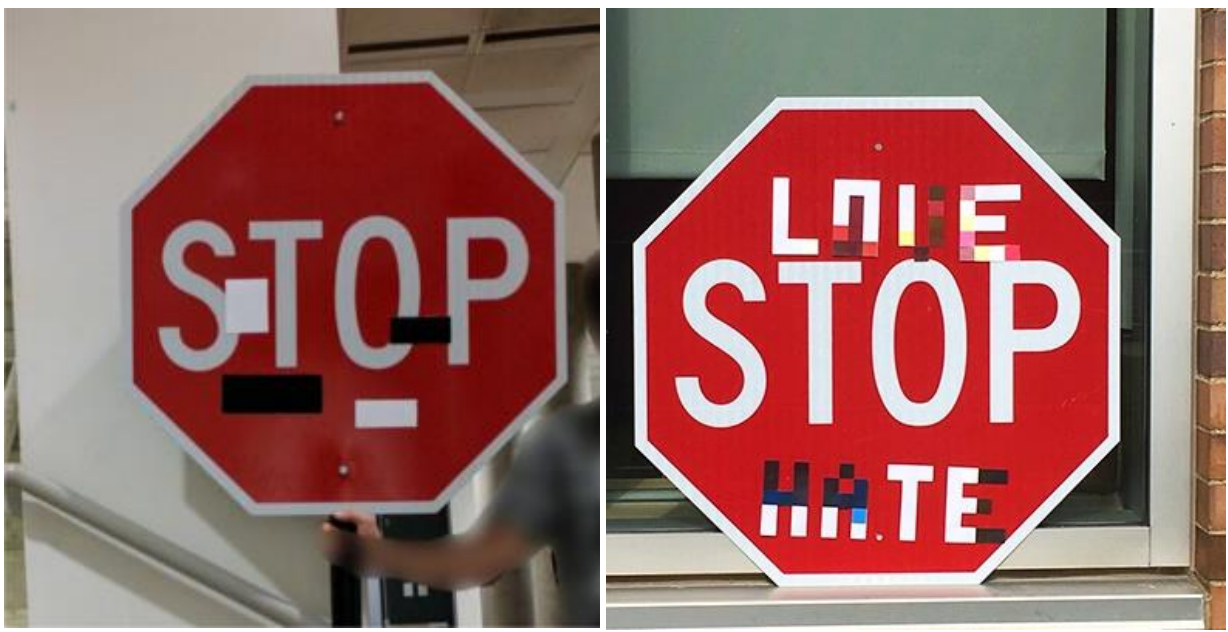

Fig 2. On the left, a real stop sign marked up with adversarial decals that cause it to be coded as a yield sign. On the right, an adversarial decal is disguised to look like normal human graffiti (Eykholt et al., 2017).

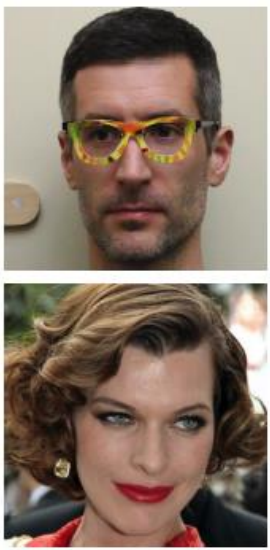

Fig. 3. A pattern which can be printed on a set of "adversarial glasses" was discovered by Sharif et al. (2016) using white-box methods (i.e., with god's-eye access to model parameters). Wearing these glasses in front of a DNN trained on facial recognition caused it to mis-classify a 40-year-old white male 87% of the time as being the actress Milla Jovovich.

**2. Recent Re-examinations of the Adversarial Attacks**

Adversarial examples shook the nascent deep learning world upon their discovery, leading to an intense wave of research. These investigations showed that nearly every intuition researchers had about adversarial examples was wrong (for a review see Yuan et al., 2019). Researchers first expected that their effects would be idiosyncratic and tailored to a particular model, and they would apply only to particular models to which the attackers had "white box" access (i.e., they could access all of the model's parameters and training set). To their surprise, however, the modelers found that adversarial examples created for one deep network transferred well to others treated as black boxes—and even transferred with the same incorrect labels. Another idea was that they could be overcome using simple "regularization" techniques such as de-noising or rotation. While this worked on some adversarial techniques—especially on early "perturbed images", such as the panda-gibbon example—it did not work on all of them (Xu et al., 2017). In particular, other methods for creating adversarial examples focused on generating nonsense images or adding meaningless swatches to normal images, both of which could cause the image to be (mis)categorized with high confidence by DNN, but which appeared nonsensical to humans (Nguyen et al., 2015). These so-called "rubbish" images proved to be a more persistent source of adversarial examples that could be deployed in real-world settings, such as the stop-to-yield decals or Milla-Jovovich glasses discussed above.

Begrudgingly accepting the resilience of the phenomenon, researchers then began casting about for explanations—but again, most intuitions were incorrect. The first idea explored was that the nonlinearity of common deep-learning architectures which was responsible for much of their power was also responsible for their vulnerability to adversarials (Goodfellow et al., 2015). One of the key differences between state-of-the-art deep learning networks and earlier neural networks is that they use several different types of nodes with different activation functions, some of which compute linear combinations of their inputs, and others which compute a nonlinear function of their inputs (for a full explanation and review of the technical significance of these features, see Buckner, 2019). In particular, they often use a function called max-pooling, which was intended to simulate the nonlinear response profiles of "complex neurons" in the mammalian visual cortex which seem to allow for the detection of abstract features across forms of perceptual variation such as location in the visual field, rotation, pose, and so on (Fukushima, 1979). However, it was eventually discovered that the problem was actually that the models can still be *too linear* in their classification of points that are far away in data space from the examples on which they are trained (perhaps as a result of the "curse of dimensionality" wrought by the oddness of distance-metrics in high-dimensional space—Domingos, 2012). These models combine pooling with more linear operations, and small differences from perturbation or rubbish swatches can accumulate through many layers of ReLU processing to produce highly confident misclassifications when passed through the final classification layer. However, these linear operations were inspired by the behavior of "simple cells" in mammalian visual cortex, and it is not clear how to get the models to categorize exemplars correctly or model human perceptual cortex without combining linear and non-linear processing in this way (DiCarlo et al., 2012; Hubel & Wiesel, 1967).

**3. Are These Verdicts on Adversarial Examples Bugs, or Features?**

Most recently, accepted wisdom about adversarial examples has again been shaken again by a series of discoveries suggesting that vulnerability to them might not be so undesirable, after all. One surprising paper devised a method to create perturbed images that could fool time-limited humans (Elsayed et al., 2018, Fig 4), suggesting that even human perception may be vulnerable to some adversarial attacks. Another recent study by Zhou & Firestone (2019) presented human subjects with a series of rubbish images generated by a wide variety of different methods and in a wide variety of different experimental setups (Fig. 5). When human subjects were given a choice of which labels they thought computer systems were most likely to assign to the images, they were able to guess the right label at rates well above chance for a most of the adversarial images tested. Because the human subjects could tell how the computer models were likely to label the images, the authors hypothesize that deep networks really are modeling perceptual similarity judgments in humans; the problem is just that these models lack the resources to tell the difference between what something *looks like* and what it *really is*. This may be what we should have expected, given the ability to draw this distinction is not something that was built into their architecture, nor is it a task on which they were trained.

Challenging accepted wisdom even further, however, a study by Ilyas et al. (2019) delivered the most surprising findings yet. Their empirical study found two things: 1) that the classification performance of systems which were trained exclusively on adversarial images generalized well to novel natural images; and 2) when non-robust features were removed from training sets, deep nets were no longer so vulnerable to novel adversarial examples, but their accuracy in classifying natural images was also diminished. They concluded from these findings that deep nets were discovering highly-predictive features in adversarial examples that generalized well to novel real-world data—which finally provides an explanation for the otherwise-curious transferability of adversarial examples from one model to others with different architectures and training sets. They are learned by these other models because these features bear predictively-useful information that is reliably present in real-world input data. An innovative form of commentary and rebuttal concerning these findings was published by several research groups in the online computer science journal *Distill*; this exercise found that even those who were initially skeptical of Ilyas et al.'s results were not only unable to refute their empirical argument, in the end they replicated and largely extended their findings (Engstrom et al., 2019).

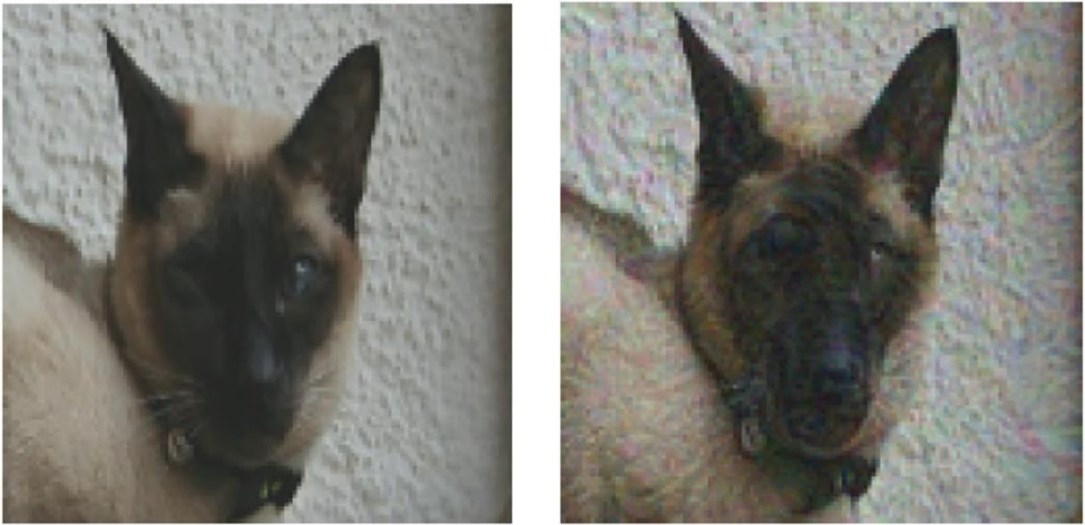

Figure 4. A perturbed image that can purportedly fool human subjects, with the original image of a cat on the left, and the perturbed image (often classified as a dog) on the right (from Elsayed et al., 2018).

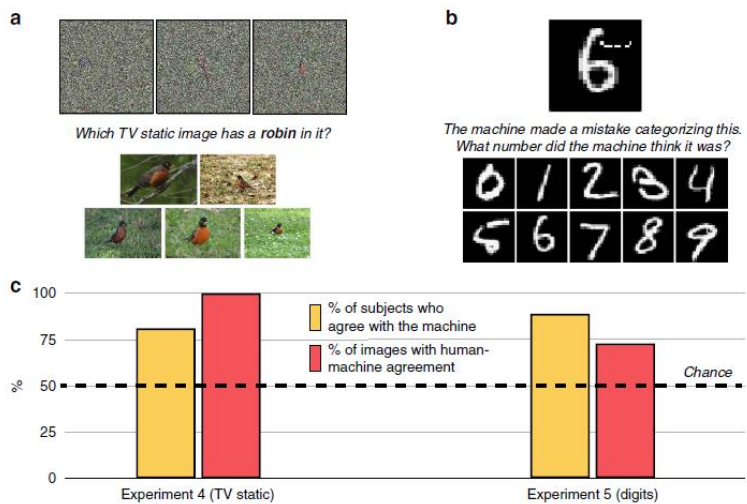

Figure 5. Two of the experimental designs from Zhou & Firestone (2019). For these two adversarial example generation methods, human subjects agreed with machine-assigned labels across all images well above chance, and a significant number of specific adversarial examples tested exhibited above-chance levels of human-machine agreement.

Collectively, these results leave us with the unavoidable conclusion: however strange it may sound, the features which are responsible for deep learning models' sensitivity to adversarial features are not just the result of overfitting idiosyncracies or noise in the structure of a specific training set. They are "non-robust", in the sense that humans for whatever reason cannot recognize them—perhaps due to our lower perceptual acuity, or cognitive inability to grasp "high-frequency features" that are distributed throughout the image; but they are useful tools in predicting labels for natural data, as confirmed by standard methods of assessing feature validity in machine learning like cross-validation. However, machine learning researchers are now posing a host of questions that these methods are ill-suited to answer. Are these features "real" aspects of the objective world (Ilyas et al. call them part of a category's "intrinsic data geometry")? If humans could recognize them, would we use them in our own categorizations?

Is this starting to sound less like computer science and more like philosophy to anyone else, yet?

4. **A Philosophical Interlude: The Deeper Riddle of Induction**

If the non-philosophers in the audience will permit a brief digression, I suggest that here, machine learning researchers are confronting foundational questions in epistemology and philosophy of science that are akin to those raised by Nelson Goodman in what he called his "new riddle of induction" (1983). Goodman challenged us to explain why scientific generalizations like "emeralds are green" can be confirmed by their instances (i.e., finding a green emerald), but nearby generalizations like "emeralds are grue"—where 'grue' is defined as "green before time $t$ or blue after time $t$"—cannot. The moral of the story, Goodman proposed, is that 'green' is predicate which is "projectible" onto future cases, but 'grue' is not. This, however, merely gave a name to the problem's solution, for unless we have an independent account of projectibility, we have no way to decide whether a predicate is suitable for inductions. Goodman then explored several different ways to cash out this idea of projectibility: perhaps the problem was that the definition of 'grue' makes reference to specific spatiotemporal features, or that it is defined in terms of other predicates. All of these attempts failed, however, because the supposedly projectible predicates like green could also be defined in these ways (e.g., we could as well define "green" as "grue before time $t$, but bleen afterwards"). Goodman concluded somewhat pessimistically that the only reason to favor 'green' over 'grue' is its entrenchment in our classificatory and justificatory practices, which for all we know might reflect an arbitrary historical contingency.

Here, Quine (1969) famously appealed to evolution to soothe our troubled nerves. He suggested that the entrenchment of certain predicates in scientific practice was not an accident: some predicates jump out at us as natural candidates for inductions because evolution has shaped our perceptual and cognitive faculties to respond to some properties and not others, because some tracking those properties reliably helped our ancestors survive and reproduce. In other words, we find the mutual similarity of green objects to be salient because it was useful to our ancestors to recognize green things, but not grue ones. This provides us—at least, those of us contentedly cruising along in Neurath's boat with the evolutionary biologists—with some defeasible justification to trust that our default inclinations towards certain "natural" predicates will not lead us astray. Many other influential theorists followed Quine here; Putnam, Millikan, and Boyd all gave natural properties pride of place in their metaphysics and philosophy of science, and some of their most influential work explores how we manage to conduct successful scientific discourse using them before we know whether it will pay off (Boyd, 1999; Millikan, 1999; Putnam, 1975).

However, we might now wonder whether adversarial examples reveal an alternative fork in the road that we might have taken before setting down this more familiar Quinean path. The relevance of machine learning to these foundational questions about the nature of scientific reasoning has been recognized before (Harman & Kulkarni, 2012; Thagard, 1990), but the recent discovery of adversarial examples forces us to reconsider them afresh. The features detected by deep networks which are responsible for their susceptibility to adversarial examples are said to be "non-robust", which in machine learning lingo means they are "incomprehensible to humans" (Ilyas et al. 2019)—and hence must be regarded as non-natural in the Quinean sense. However, we might recall that Quine was at his core a pragmatist, and his reassurances to trust our innate sense of similarity was only a pragmatic justification for relying on default perceived similarities as empirically fecund. Quine's arguments did not establish that other non-natural predicates might not also be projectible into the future, in the Goodmanian sense; and in fact he suggested that our reliance on a naïve sense of naturalness was only a waystation in the development of an adolescent science anyway, and that mature sciences would eventually "slough off the muddy old notion of kind or similarity piecemeal, a vestige here and a vestige there" until all that mattered for ontology was which predicates featured in the most highly-confirmed and empirically successful scientific theory for some domain (1969, p53).

What, then, if deep learning networks have revealed to us a different starting point for this scientific expedition—guided not by a native sense of similarity constrained by the tenuous course of hominid evolution and perceptual failings, but rather by the sharper and more informed eyes of artificially-engineered deep learning agents? Could sensitivity to these features explain how deep learning agents are able to defeat humans so handily on tasks that seem beyond our ken—such as interpreting fMRI data, predicting protein folds, or seeing strategies in Go which span the entire board? If scientific investigation would become more productive—allowing more prediction, control, and other scientific goods—by following these high-frequency bread crumbs, then Quineans of pragmatic stripe should likely embrace this alternative path towards scientific progress. In normative epistemology, however, we will eventually have to confront a roadblock which has been laid elsewhere in philosophy of science regarding the nature of explanation (Khalifa, 2013; Potochnik, 2011). Assuming that humans are never able to see and intuitively grasp non-natural features ourselves—through the use of augmented reality headsets, for example—then it is unlikely that explanations phrased in terms of non-natural features should ever produce in us that satisfying epistemic feeling of understanding that some regard as essential for successful explanation (Arango-Muñoz, 2014).

Setting aside the muffled hurrah which might be heard from the quantum physicists who agreed with Bohr that scientists should stop telling God what to do with the universe: can so much future science be regarded as a progress, if its achievements do not provide humans with any increased sense of understanding? And what are these non-robust but useful features, anyway? Should science rely on them? And how could we decide? These are the questions we explore in the final sections.

**5. The need for a theory of artifacts in deep learning**

Suppose, for the sake of argument, that there are scientific disciplines in which progress may depend in some crucial way upon detecting or modeling predictive but human-inscrutable features. To ground the discussion in a specific example, let us return to protein folding. For many years in the philosophy of science, protein folding was regarded as paradigm evidence for "emergent" properties—properties which only appear at higher levels of investigation, and which cannot be reduced to patterns in lower level details (Humphreys, 2016). The worry here was that the interactions amongst amino acids in a protein chain were so complex and nonlinear that humans would never be able to reduce global folding principles to chemistry and physics (Theurer, 2014). This was often discussed under the heading of "Levinthal's paradox", after Levinthal speculated that there were so many degrees of freedom for in an unfolded polypeptide chain (he estimated $10^{143}$—which we might compare to the massive branching factor of Go or the dimensionality of image classification) that it was a mystery how only a small number of structures prove stable at the end of such a rapid folding process in the first place (Honig, 1999; Zwanzig et al., 1992). Instead, human scientists relied on a series of analytical "energy landscape" or "force field" models that could predict stability of final fold configurations with some degree of success. These principles are intuitive and elegant once understood, but their elements cannot be reduced to the components of the chain in any straightforward manner, and there seemed to be stark upper limits on their prediction accuracy. By contrast, AlphaFold on its first entry in the CASP protein-folding competition was able to beat state-of-the-art analytical models on 40 out of 43 of the test proteins, and achieve an unprecedented 15% jump in accuracy across the full test set (Senior et al., 2020).

Subsequent work has suggested that the ability of deep learning neural networks to so successfully predict final fold configurations may depend upon the identification of complex "interaction fingerprints" which are distributed across the full polypeptide chain (Gainza et al., 2019). We might speculate that these interaction fingerprints are like the sorts of non-robust features that cause image classifying networks to be susceptible to adversarial attacks, in that they are highly-complex, distributed across the entire input, predictively useful, and not amenable to human understanding. In other words, the same features of DNNs that make them susceptible to adversarial examples might enable them to predict protein folds better than seemingly more transparent analytical models. Suppose this is all the case, for the sake of argument: should protein science rely on these interaction fingerprints going forward?

The answer, I will suggest, is a big: it depends. Before saying what it depends upon, we need to clear out a bit of conceptual space. Many discussions about adversarial examples and non-robust features have become trapped by a simple dichotomy between **signal** and **noise**. The initial assumption was that DNNs were only vulnerable to adversarial attacks because they overfit noise in the data set. As I have mentioned above, this conclusion is at least partially rebutted by the results of Ilyas et al. (2019); at least some significant portion of the time, high-frequency features are not merely aggregations of random noise, but rather reflect predictively useful patterns in natural data. This finding might tempt us to swing to the other side of the dichotomy, and assume that if these features are predictively useful, they must reflect real "signal" about the categories we want to track. This, however, would also be too hasty, for there remains a third conceptual category from signal processing theory that has been little-discussed in this context: that of **artifact**.

Artifacts are not simply random noise, because like signal they are often predictively useful; but neither are they objective features of the input that we necessarily want to track. Let us consider some simple examples to get the intuitions flowing. Lens flares in photography are a commonly-encountered type of a visual artifact. Lens flares can contain predictively useful information about real objects depicted in the photograph; they provide useful information about the location and intensity of light sources in the scene depicted. However, they are not observer-independent features of these scenes, nor are most of their visual properties directly isomorphic with properties of those light sources. The visual qualities of a lens flare are as much a product of the lens diameter and filtering as they are of the light sources that cause them to appear on a photograph. Lens flares do not fool us, because we are familiar with the phenomena and its causes; we already have well-developed formal and informal theories of such artifacts in photography.

Predictively-useful artifacts can even occur in human cognition. Consider, for example, the visual migraine aura. This is a distinctive and stereotyped perceptual experience, beginning with a blind spot in the center of one's vision, which gradually progresses outward in a semicircular jagged "scintillating scotoma" through the sufferer's field of vision as visual acuity gradually recovers behind it (Fig 5). This process takes about 30-45 minutes, after which the individual's visual acuity is largely restored. During my postdoctoral fellowship in Germany, there was a particular forest on my favorite cycling route where the pattern of trees would reliably induce a visual aura, likely due to the frequency and intensity of bright sunlight flickering through the canopy (bright, flickering light being a common migraine trigger). These were the only times I experienced visual migraines during my time in Germany. In this context, the visual aura was thus perfectly predictive of cycling through this particular forest on a sunny day. However, we would obviously not want to call the visual aura a "real feature" of this forest; it was instead an artifact caused by statistical properties of this forest's illumination (interacting perhaps with dehydration and physical exertion) which produced a chain reaction of spreading synaptic depression areas of the visual cortex responsible for rendering visual experience (Ayata, 2010). They are, in other words, "painted on" visual perception of the environment by cortical processing malfunctions which are induced only by a limited range of environmental circumstances.

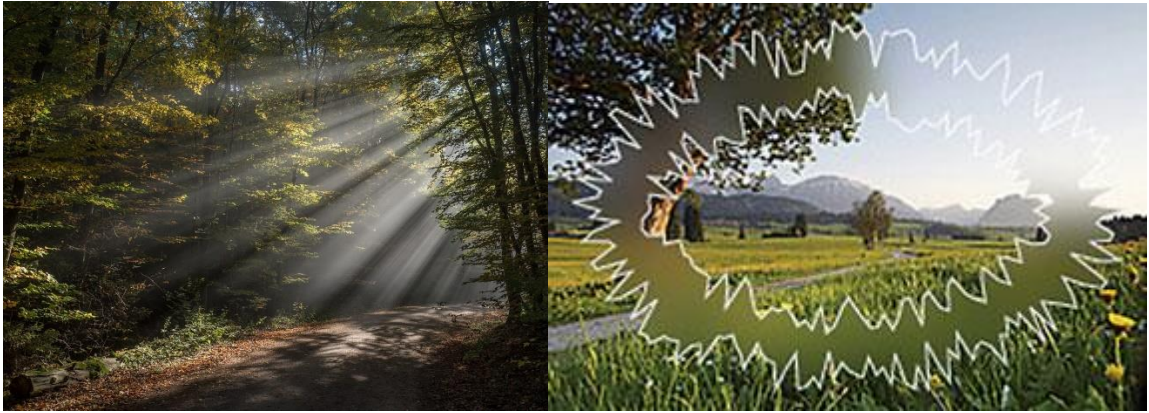

Figure 5. On the left, light pattern streaming through a forest from Bruderwald Forest in Bamberg, Germany (credit Reinhold Möller, available on a Creative Commons ShareAlike4.0 license.) On the right, an artist's depiction of a visual migraine aura, reproduced from Banyas, 2015)

To review another example from a discipline more closely-related to machine learning, consider the Gibbs Phenomenon in signal processing theory (Fig 6). The Gibbs phenomenon is caused by an "overshoot" in the Fourier series of an input signal when the target function approaches a jump discontinuity. To explain, a Fourier series for some differentiable function is a decomposition of that signal into a weighted summation of sinusoid waves with different amplitudes and frequencies; approximating that function with a Fourier series can help simplify the solution to a variety of scientific problems. As the number of sinusoids in the summation increases, the Fourier series can come very close to approximating a wide variety of differentiable functions. However, jump discontinuities present an enduring challenge, because adding more sinusoids does not eliminate the overshoot. At the limit, the overshoot disappears; but adding any finite number of sinusoids to the summation only "squashes" the overshoot closer and closer to the jump discontinuity. Like Ilyas et al.'s non-robust features, this overshoot is useful in the sense that it contains highly predictive information about the location of the jump discontinuity in the target signal; bit is also misleading about the value of the target for the duration of the overshoot. It is thus a useful artifact, and whether or not we should deploy it in our data analysis depends upon whether the value of the signal at the overshoot is relevant to our purposes and how we interpret it.[1]  Like the migraine auras, the overshoot is a product of a possibly undesirable interaction between statistical properties of the input and the processing method used to simplify that input for the purpose of classification and decision-making.

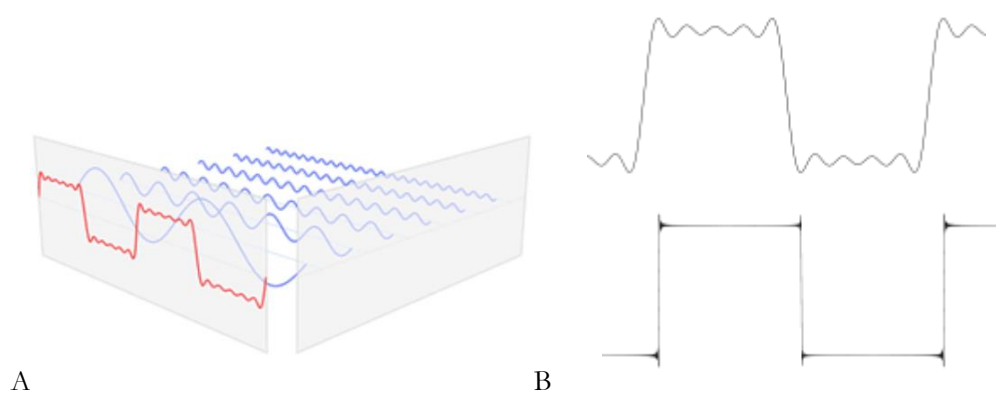

A                                              B

Fig 5.  A periodic signal function can be approximated by weighted summation of sinusoids.  Figure 5A shows the approximation of a periodic step function (in red) by a summed series of weighted sinusoids (in blue).   Figure 5b (top) shows how the summation "overshoots" the peak of the signal before settling into a stable echo; and (bottom) shows that while adding more elements to the summation can "squash" the overshoot closer to the point of the jump discontinuity, the overshoot is never fully eliminated with a finite number of sinusoids.

It is possible—perhaps even likely—that the late-stage, transformed scalar signals at the end of a deep neural network's processing hierarchy could, in certain conditions, similarly produce predictively-useful artifacts. We already know that such artifacts can be found in the products of Generative Networks, an architecture which deploys near-inverses of the typical DCNNs (the operations are often called "deconvolution" and "unpooling") to reverse the operations of a discriminative deep convolutional network and produce the photorealistic artificial stimuli and "deepfakes" that have so captivated the popular press (Goodfellow, 2016). The initial versions of this technology produced undesirable errors like "checkerboard artifacts" (Odena et al., 2016, Fig 6).  Checkerboard artifacts are easily recognizable and undesirable features of these networks' output, even though the checkboards can contain useful information about the typical locations of object edges depicted in the network's output.  As with lens flares, we quickly developed a theory of the origins of checkerboard artifacts—they are caused by an interaction between the stride length chosen for the deconvolutional operation and statistical features of the data input. Once the source of the artifact was identified, we could develop countermeasures to lessen or eliminate them.

---

[1] I thank Andy Terrell for suggesting this example.

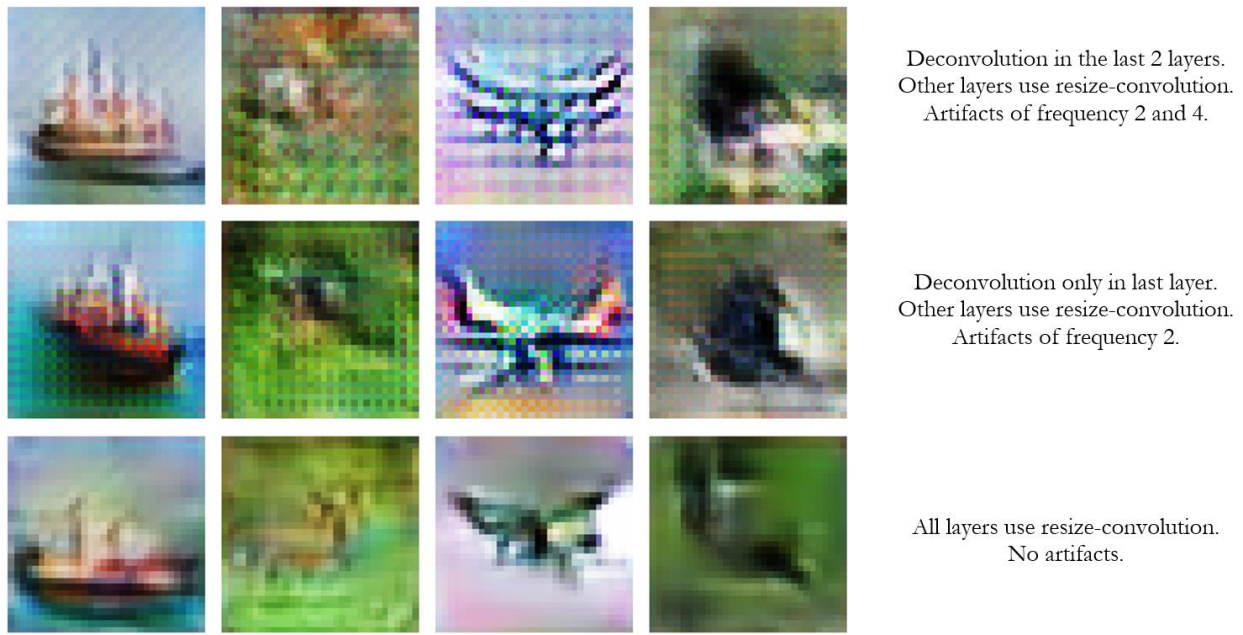

Figure 6. Examples of checkerboard artifacts produced by image deconvolution with and without corrective measures. Specifically, Odena et al. advocate for an alternative deconvolution method—"resize-convolution"—that reshapes images to fit kernels before attempting to reverse the convolution operation (figure reproduced from Odena et al. 2016).

Should we similarly conclude, then, that the non-robust but predictively features discovered by Ilyas et al. (2019) which look like errors to human perception should be dismissed as artifacts and eliminated from future scientific investigations? This conclusion could also be too hasty, for at least two reasons. First, we need to balance the possibility that non-robust features are deep learning artifacts against the possibility that they are real features detectable only via "alien" perception or cognition. We do not yet know what ground truth is for cases at the frontier of scientific progress, so we cannot conclude that they are artifacts simply because humans cannot recognize them. Let us return to color vision for some more "intuition pumps" that illustrate this latter possibility. Many animals (and some humans) are tetrachromats, which means they have extra color-detecting cells in their retinas. This allows them to detect stark differences in perceived colors that would be virtually invisible to a trichromat (such as your average human). These additional color contrasts might be vitally important in understanding the allure of a bird's mating display, which can hinge on presenting an area of plumage which appears highly salient to tetrachromats, but inscrutably bland to trichromats (Fig 7). In this sense, we may call the additional colors perceived by tetrachromats real features of this plumage, even if they are not scrutable to trichromats. This observation complicates the intuitive distinction between "intrinsic features of the data geometry" and features which are observer-dependent and "painted upon" the input by the processing architecture.

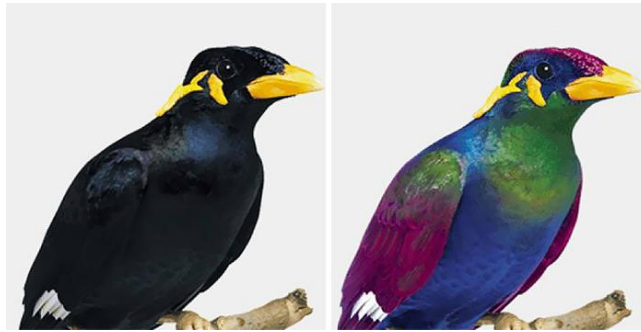

Figure 7. An artist's rendition of how a tetrachromat bird may perceive (what appears to trichromats as) blandly dark plumage. (Of course, we can only guess what their color experience is really like; this is the point of the claim that the tetrachromat's color experience is inscrutable to trichromats.)

The second and related reason we cannot so quickly dismiss Ilyas et al.'s non-robust but useful features as artifacts is that the concept of "artifact" is surprisingly difficult to define. Deciding what is or is not an artifact depends upon the function of the feature in our investigation, and so may be domain- and even purpose-specific. In information science, artifacts are often defined as features of images or signals which are introduced by the processing methodology but which are "unintended" or "undesired". We can even return to lens flares to illustrate this problem; if you are J.J. Abrams directing the *Star Trek* reboot movie, the (numerous) lens flares the footage contains—one fan counted 721 in the first reboot film—may be regarded as intended features of the scene. When confronted with the sheer number of these flares in the movies, Abrams remarked that "the idea [was] that the future that they were in was so bright that it couldn't be contained and it just sort of broke through" (Acuna, 2015). Love them or hate them, these flares are not artifacts; Abrams intended them to convey aesthetic or emotional aspects of these scenes to the film's audiences (he purportedly even added some using CGI in post-processing). We can similarly imagine scientific applications of the Gibbs phenomenon in which the overshoots are desireable and intended, accentuating useful information about the location of a jump discontinuity. Knowing how to interpret the aspect of the signal in all cases proves crucial; for example, if someone misinterpreted a lens flare in a *Star Trek* scene as depicting a phaser beam, or the peak value of a Gibbs phenomenon as a proxy for the magnitude of the original signal, then things would have gone awry.

Deciding whether something counts as an artifact in human perception may be similarly troubled. Both of these complications—being observer- and function-relative—can be found again in simple color perception in humans. We need only return to the notorious online debate over #thedress to illustrate this point (Fig 8). First, whether the dress is perceived as blue or gold depends upon assumptions about lighting conditions, because our perceptual processing contains sophisticated mechanisms to maintain perceived color constancy in the face of very different illumination conditions (Toscani et al., 2017). The wavelength of light coming from what is perceived as blue in one set of conditions and gold in another is identical, though the experiences are very different. Such constrasts caused Early Modern philosophers like John Locke to draw a distinction between primary qualities like number and shape (thought to be objective and detectible from multiple sensory modalities like sight and touch), and secondary qualities like color which (normally) seem bound to a single modality and projected on experience by the mind (Locke, 1841). Many people responded to the original image of the dress by demanding to know what color the dress is "really"…but in cases where the images are artificially generated as in Fig. 8, there is no clear answer to this question. This gets to the second issue of function-dependence of artifact-hood: the right answer—and whether one even exists—depends on origins of the data and why we want to make discriminations about them in the first case. This is why adversarial examples are so confounding in the present debate: they are (usually) artificially-generated, and though humans reliably perceive them to contain a certain type of object, there is (usually) no ground truth as to what objects they "really" depict (cf. Fig 8). They have no purpose except to cause divergence between the verdicts between humans and DNNs, leaving us with scant resources to draw upon when we come into doubt as to which verdict is actually correct.

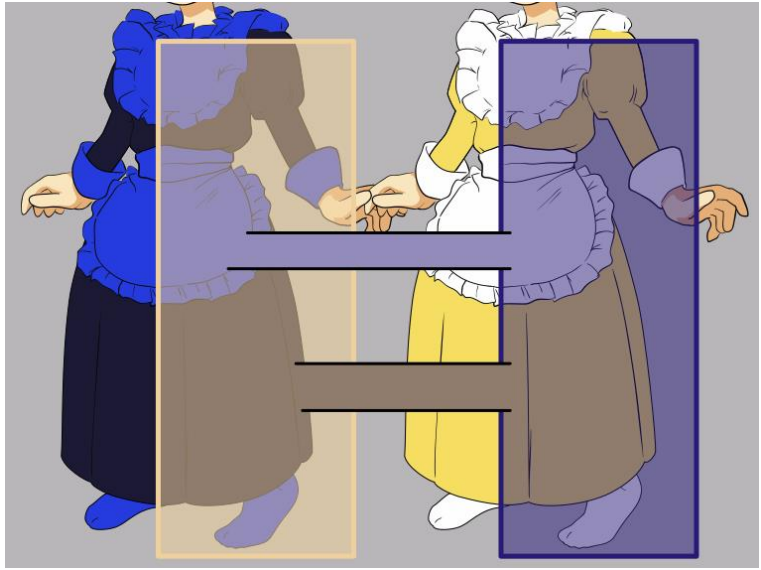

Figure 8. Figure design by Kasuga~jawiki; "The dress" modification by Jahobr, available under a GNU free document license 1.2 from Wikimedia Commons.

So, setting aside the problem of adversarial examples, and focusing on the role of useful but inscrutable features in future scientific investigations—should we use them or not? Immanuel Kant famously extended Locke's skepticism about secondary qualities to primary qualities like shape and number, worrying that nothing we perceived could be guaranteed independent of processing constraints imposed by our cognitive architecture (Smith, 2011). When we consider the limits of scientific investigation—in quantum physics, protein folding, or climate science—we must take the possibility of exceeding those limits seriously (Boudry et al., 2020). There is little reason to suppose that evolution would have endowed us with cognitive machinery adequate to prediction and control of natural phenomena in these highly-complex domains which were inaccessible to our evolutionary ancestors, and already ample evidence that our minds are ill-suited to the task. Neither, however, can we set off blindly trusting the verdicts of DNNs trafficking in features we find inscrutable, given the danger that they are artifacts.

**6. Towards a solution: Elements of a theory of artifacts for deep learning**

I suggest three interlocking strategies to advance the current debate over adversarial examples and human-inscrutable science. First and most generally, we need a taxonomy and corresponding etiological theory for artifacts that arise in deep learning neural networks. The first step might be to taxonomize different types of useful but non-robust features. Some of this has already begun in response to Ilyas et al.'s original finding; Gabriel Goh proposed at least two different kinds of non-robust features, "ensembles" and "containments" (Goh, 2019, Fig 9). Goh defined ensembles as collections of non-robust and non-useful features which, if sufficiently uncorrelated, could be combined into a single useful, robust feature. These might be things we would want to retain in mature science—the sort of thing we might call a "texture", if we could recognize it. Containments, on the other hand, are interpolations of a useful, robust feature and a non-robust, useless feature. Because containments could always be replaced with a more useful feature that predicted the same outcomes, there seems little reason why we would want to retain them in a robust science.

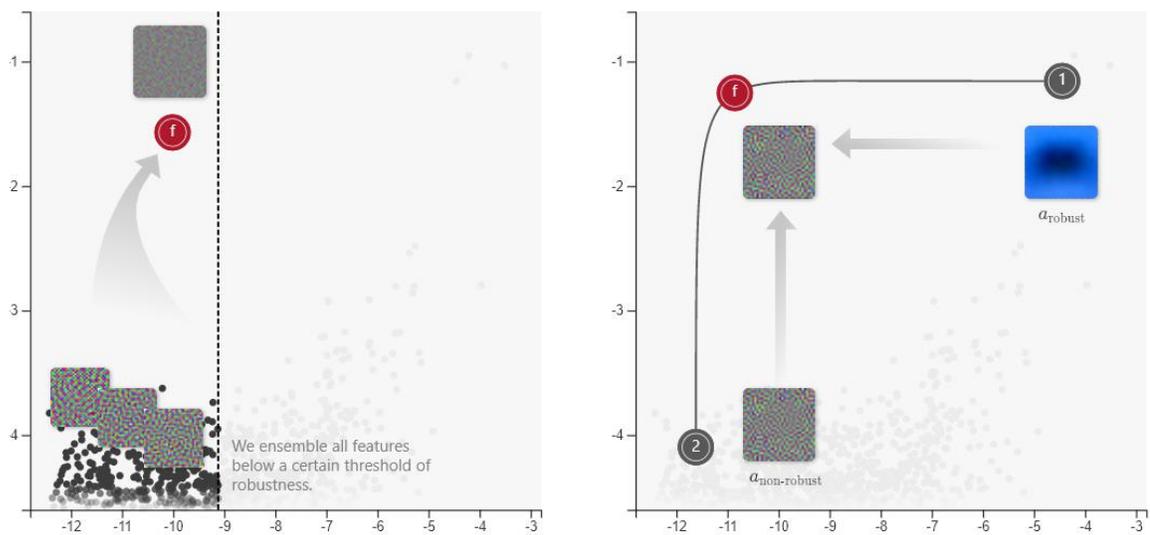

Figure 9. Example of an ensemble feature on the left, and a containment feature on the right, reproduced from Goh (2019).

In tandem with developing a viable taxonomy of such features, we should try to uncover the etiologies of each sub-type—that is, to discover which architecture or hyperparameter choices tend to produce which types of features. There are many hyperparameter choices that go into the construction of these networks—number of nodes in each layer, number of layers, types of activation functions, regularization methods, and so on. As we discovered with checkerboard artifacts being produced by the stride length of deconvolution, we are likely to find that certain features are produced by certain types of parameter choices. This would be a first step towards mitigating the appearance of these features in our outputs, should we decide that they are undesirable artifacts.

Discovering the etiologies for different types of features would also help with the second strategy for detecting artifacts in deep learning—the method of triangulation. Many other sciences also have to make do with methods of investigation which are not fully trustworthy and which cannot be calibrated against gold standard data or independent accounts of ground truth. In sociology, for example, there are a variety of different survey and investigative methods, but none of which can be regarded as fully trustworthy. A standard method in this discipline is the strategy of triangulation—authors try to use as many qualitatively different methods to ask the same question as possible, and regard an answer to be more likely as true than as an artifact of a particular method if it arises independently from multiple independent sources of evidence (Denzin, 2017).[2] In other words, we may want to try many different machine learning methods with different hyperparameters on the same data; if the same type of feature reliably appears in the same way on many different methods, it may be less likely to be an artifact. A complication of this approach, however, is that without knowing which of the hyperparameters produces which type of feature, we won't know what aspects of the architecture to vary when looking for independent confirmation. For this reason, this second aspect of developing a theory of artifacts for deep learning can only be conducted in concert with the other two.

Finally, a default purpose we have for conducting scientific investigations is the desire to control the forces we study, and for this reason we may rely upon the possibility of effective causal interventions as an arbiter of artifact-hood scientific applications of deep learning. Machine learning researcher Zachary Lipton has recently expressed concern about what he calls the "meta-meta problem" of deep learning research—too many researchers exhibit a tendency to treat every application as a mere prediction problem, when really we have many other purposes in mind which are not served by mere prediction. A concern for causal potency is a venerable idea in philosophy, traceable back to the "Eleatic Principle" from Plato's dialogue *The Sophist* which is often paraphrased as "to be is to have causal power." Notably, this principle would have easily dealt with the question of whether the migraine aura is a real feature of the forest; attempting to intervene on the scintillating scotoma would be fruitless in the attempt to cause any changes in the forest. Even if we found a way to intervene directly on the migraine aura—by taking medication, for example, or increasing the amount of magnesium in our diet—this would not change any other properties of the forest. To return to the example of protein folding, we often attempt to predict protein folds not simply for their own sake, but for a variety of practical purposes: trying to find new ways to diagnose Alzheimer's, for example, or developing new drugs. If we have no way to predict the result of interventions on an interaction fingerprint, then they may not be useful for these purposes. However, if deep learning could also help us intervene on these fingerprints to intentionally produce certain folds—by telling us the necessary concentrations of chaperone molecules needed to produce the fold, for example—then this would provide strong evidence that the interaction fingerprint is not a mere artifact. Even secondary qualities like color may satisfy this requirement in the right context; consider the importance of albedo in models of climate change, where color differences can dramatically change how much heat a surface absorbs from the sun (Winton, 2006). However, regarding causal manipulability as a necessary condition might be too strong; some sciences might make do with a weaker notion of "construct validity" that established the robustness of the pattern across a variety of situations without grounding a meaningful sense of intervention (Cronbach & Meehl, 1956; Sullivan, 2016).

---

[2] I thank Thomas Lindner for suggesting the example of triangulation in sociology.

Unfortunately, modeling the difference between causation and correlation has been regarded as something of a characteristic weak spot for deep learning, and so we may need to supplement it here with alternative techniques (Pearl, 2019). In philosophy, explicitly interventionist theories of causation may provide some useful conceptual guidance (Woodward, 2003), and in computer science Causal Bayes Nets could be deployed in tandem with deep learning networks to develop experiments which would help discern which features are merely predictive and which ones also have causal potency (Pearl, 2009). Researchers are also helping to develop methods of causal inference within the deep learning paradigm (Battaglia et al., 2018; Dasgupta et al., 2019). The need for a theory of artifacts provides renewed motivation for integrating causal reasoning into the methodology of deep learning, and whichever approach is preferred, provides it with another important test application and links to other important areas of machine learning research. And finally, a focus on causal interventions can help soften the blow of seemingly-inscrutable scientific progress. Philosophers of science have distinguished a variety of different dimensions of explanatory power, only one of which is "cognitive salience" to humans. Many others—such as non-sensitivity, precision, factual accuracy, and degree of integration with background theory—may be satisfied by causally-useful-but-human-inscrutable features and traded off against losses in cognitive salience, providing us with a principled way to decide when they ground explanations even if humans cannot understand them (Ylikoski & Kuorikoski, 2010).

Sharif, M., Bhagavatula, S., Bauer, L., & Reiter, M. K. (2016). Accessorize to a Crime: Real and Stealthy Attacks on State-of-the-Art Face Recognition. *Proceedings of the 2016 ACM SIGSAC Conference on Computer and Communications Security*, 1528–1540. https://doi.org/10.1145/2976749.2978392

Silver, D., Schrittwieser, J., Simonyan, K., Antonoglou, I., Huang, A., Guez, A., Hubert, T., Baker, L., Lai, M., & Bolton, A. (2017). Mastering the game of go without human knowledge. *Nature*, *550*(7676), 354.

Smith, N. K. (2011). *Immanuel Kant's critique of pure reason*. Read Books Ltd.

Sullivan, J. A. (2016). Construct Stabilization and the Unity of the Mind-Brain Sciences. *Philosophy of Science*, *83*, 662–673.

Thagard, P. (1990). Philosophy and machine learning. *Canadian Journal of Philosophy*, *20*(2), 261–276.

Theurer, K. L. (2014). Complexity-based Theories of Emergence: Criticisms and Constraints. *International Studies in the Philosophy of Science*, *28*(3), 277–301. https://doi.org/10.1080/02698595.2014.953342

Toscani, M., Gegenfurtner, K. R., & Doerschner, K. (2017). Differences in illumination estimation in #thedress. *Journal of Vision*, *17*(1), 22–22. https://doi.org/10.1167/17.1.22

Winton, M. (2006). Amplified Arctic climate change: What does surface albedo feedback have to do with it? *Geophysical Research Letters*, *33*(3).

Woodward, J. (2003). *Making things happen: A theory of causal explanation*. Oxford University Press. http://books.google.com/books?id=LrAbrrj5te8C&pgis=1

Xu, W., Evans, D., & Qi, Y. (2017). Feature squeezing: Detecting adversarial examples in deep neural networks. *ArXiv Preprint ArXiv:1704.01155*.

Yamins, D. L., & DiCarlo, J. J. (2016). Using goal-driven deep learning models to understand sensory cortex. *Nature Neuroscience*, *19*(3), 356.

Ylikoski, P., & Kuorikoski, J. (2010). Dissecting explanatory power. *Philosophical Studies*, *148*(2), 201–219.

Yuan, X., He, P., Zhu, Q., & Li, X. (2019). Adversarial examples: Attacks and defenses for deep learning. *IEEE Transactions on Neural Networks and Learning Systems*, *30*(9), 2805–2824.

Zhou, Z., & Firestone, C. (2019). Humans can decipher adversarial images. *Nature Communications*, *10*(1), 1334. https://doi.org/10.1038/s41467-019-08931-6

Zwanzig, R., Szabo, A., & Bagchi, B. (1992). Levinthal's paradox. *Proceedings of the National Academy of Sciences*, *89*(1), 20–22.
24